\documentclass[10pt,twocolumn]{article} 
\usepackage{simpleConference}
\usepackage{times}
\usepackage{graphicx}
\usepackage{amssymb}
\usepackage{url,hyperref}
\begin{document}

\title{PCGPT: Procedural Content Generation via Transformers}

\author{Sajad Mohaghegh, Mohammad Amin Ramezan Dehnavi, Golnoosh Abdollahinejad, and Matin Hashemi \\
Learning and Intelligent Systems Laboratory, \\ 
Department of Electrical Engineering, \\
Sharif University of Technology, Tehran, Iran.\\
}

\maketitle
\thispagestyle{empty}

\begin{abstract}
% We present a new framework for procedural content generation (PCG), the automated creation of game content such as levels, maps, items, and characters. Our framework combines offline reinforcement learning (RL) and large language models (LLMs) to generate game content in an end-to-end manner. Offline RL is a technique that learns from a pre-collected dataset of state-action-reward combinations without requiring online interaction with the environment. This enables us to train a transformer-based model that can produce the best actions for level design given the desired return, previous states, and actions. We use three different methods to formulate two-dimensional level design problems as Markov decision processes and apply them to three game domains. Our framework can generate diverse and meaningful game content that satisfies specific objectives and constraints. We also show that our framework surpasses existing PCG methods in terms of sample efficiency and generalization. PCGPT provides novel opportunities for PCG research and applications.
The paper presents the PCGPT framework, an innovative approach to procedural content generation (PCG) using offline reinforcement learning and transformer networks. PCGPT utilizes an autoregressive model based on transformers to generate game levels iteratively, addressing the challenges of traditional PCG methods such as repetitive, predictable, or inconsistent content. The framework models trajectories of actions, states, and rewards, leveraging the transformer's self-attention mechanism to capture temporal dependencies and causal relationships. The approach is evaluated in the Sokoban puzzle game, where the model predicts items that are needed with their corresponding locations. Experimental results on the game Sokoban demonstrate that PCGPT generates more complex and diverse game content. Interestingly, it achieves these results in significantly fewer steps compared to existing methods, showcasing its potential for enhancing game design and online content generation. Our model represents a new PCG paradigm which outperforms previous methods.
\end{abstract}

\section{Introduction}
Procedural content generation has become an established method for creating assets and content for video games, spanning from simple image textures to entire environments \cite{hendrikx2013procedural, shaker2016procedural}. PCG has become a promising tool for producing personalized games that adapt content to a player's preferences and optimize their gaming experience. In addition, PCG has the potential to increase the replayability, diversity, and originality of games through the creation of novel and unexpected content \cite{jialin2020yannakakis}. Nevertheless, traditional PCG methods frequently rely on predefined rules or random generation, resulting in limitations such as repetitive, predictable, or inconsistent game design or theme-related content. These techniques may also require a great deal of manual tuning to achieve the desired quality or diversity of content, and they may be incapable of generating content that adapts to the player's preferences, abilities, or emotions \cite{liu2021deep}. To surmount these limitations and generate novel content that is coherent, unique, and creative, advanced PCG techniques are required \cite{risi2020increasing}.

\begin{figure}[!t]
  \begin{center}
    \includegraphics[width=2.5in]{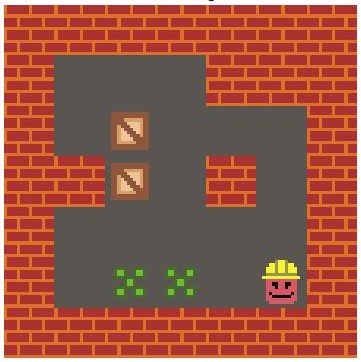}
  \end{center}

  \caption{\small A level for the puzzle game Sokoban generated by PCGPT.}
  \label{fig-label}
\end{figure}

PCG can benefit from advanced machine learning techniques that can learn complex patterns and relationships from data, such as deep learning. One of the deep learning methods that has shown great potential in many domains is deep reinforcement learning (Deep RL), which combines deep neural networks with Reinforcement Learning (RL). Deep RL has achieved remarkable results in many domains, such as game playing, where it can surpass human experts or even create novel strategies \cite{mnih2013playing}.  It can also be leveraged for PCG, which are challenging tasks that require creativity and adaptation. PCG can be modeled as a supervised learning problem, where the objective is to generate content that satisfies certain criteria, such as playability, aesthetics, or difficulty. Deep RL can learn from data \cite{khalifa2020pcgrl} or human feedback to generate novel and diverse game content, such as levels, maps, characters, rules, or mechanics.
% Creating high-quality content is a challenging task that can benefit from various deep learning techniques. One of them is reinforcement learning (RL), which is a method of learning how to act optimally in an environment by maximizing a reward signal. RL has been successful in many domains, such as playing games, where it can learn to play at a superhuman level \cite{mnih2013playing}. But RL can also be applied to game content and design, which are sometimes seen as optimization or supervised learning problems. 

Transformer model \cite{vaswani2017attention} represents a cutting-edge deep learning model known for its exceptional performance across various domains, including natural language processing (NLP) \cite{radford2018improving, devlin2018bert}, Computer Vision \cite{ViT,Swin,DE:TR}, and Time Series Analysis \cite{time_series_transformers}. It relies on a self-attention mechanism, which facilitates the capture of intricate semantic relationships within high-dimensional feature spaces. Researchers have investigated the utility of the Transformer in RL contexts characterized by sequential decision-making, as observed in gaming scenarios \cite{chen2021decision, upadhyay2019transformer, lee2022multi}. One prominent approach involves employing the Transformer to model the joint distribution of states, actions, and rewards, subsequently utilizing this model to generate optimal behavior trajectories \cite{chen2021decision}. This approach holds promise in mitigating the limitations commonly associated with traditional RL algorithms that rely on value estimation, such as issues related to high variance, instability, and sample inefficiency \cite{dulac2021challenges}.

In this research study, we employ transformer models to generate game levels by leveraging accumulated experience. We adopt a sequence modeling objective, diverging from conventional reinforcement learning techniques such as temporal difference (TD) learning, as described by Sutton \cite{sutton2018reinforcement}, which primarily focus on training a policy. This approach allows us to circumvent the necessity for bootstrapping in long-term credit assignments, a significant contributor to the well-known "deadly triad" problem that often destabilizes RL systems \cite{sutton2018reinforcement}. Moreover, our proposed method offers an additional advantage by eliminating the requirement to discount future rewards, a common practice in TD learning but one that can result in suboptimal short-term behavior. Additionally, we can leverage pre-existing transformer frameworks, widely utilized in natural language processing, known for their stability and effectiveness in training like GPT \cite{radford2018improving} and BERT \cite{devlin2018bert} models.
% This strategic choice facilitates the scalability of our approach and enables us to tap into a rich body of research on optimizing transformer models.

% In addition to their exceptional ability to represent lengthy sequences, transformers possess further advantages. Specifically, transformers are capable of conducting credit assignments directly via self-attention, unlike Bellman backups that rely on slow propagation of incentives and are vulnerable to "distractor" signals \cite{hung2019optimizing}. This feature makes transformers highly effective in scenarios where incentives are scarce or distractions are abundant. Additionally, empirical evidence indicates that adopting a transformer-based modeling approach can capture a wide distribution of behaviors, leading to improved generalization and transferability \cite{ramesh2021zero}.

In addition to their remarkable capacity for handling long sequences, transformers offer several additional advantages. Notably, transformers excel at performing credit assignments directly through self-attention mechanisms, as opposed to Bellman backups that depend on the slow propagation of rewards and are susceptible to misleading signals, often referred to as "distractors" \cite{hung2019optimizing}. This characteristic renders transformers highly effective in situations where rewards are scarce, or distractions are prevalent. Furthermore, empirical evidence suggests that employing a modeling approach based on transformers can capture a wide range of behaviors, resulting in an enhanced ability to generalize and transfer knowledge \cite{ramesh2021zero}.

We will investigate our hypothesis through the examination of offline RL. In this context, our goal is to train agents to acquire optimal behaviors using suboptimal data, which involves learning from limited, pre-existing experience. This particular task has historically posed significant challenges due to issues such as error propagation and value overestimation, as indicated by prior research \cite{levine2020offline}. Nonetheless, this task aligns naturally with the objectives of training models based on sequence modeling. By training an autoregressive model on sequences containing information about states, actions, and returns, we simplify the process of policy sampling to autoregressive generative modeling. In this way, we can determine the desired proficiency level of the policy by manually specifying the return tokens. These tokens effectively serve as prompts for the generative model, enabling us to instruct the model on which specific "skill" to focus its learning efforts.
% We test this hypothesis using offline RL, where we train agents to learn policies from suboptimal data, aiming to achieve the best possible behavior from a fixed and limited experience. This task has been difficult due to error propagation and value overestimation \cite{levine2020offline}. However, when training with a sequence modeling objective, it becomes a natural task. We train an autoregressive model on sequences of states, actions, and returns and sample policies using autoregressive generative modeling. We can also control the skill or expertise of the policy by choosing desired return tokens, which act as a cue for generation. fig. \ref{fig_overview} illustrates the high-level overview of the PCGPT framework.

\section{Related Work}
\subsection{Procedural Content Generation}
Procedural Content Generation is a computational technique used in various creative domains, notably in the context of video games and digital media, to automatically and algorithmically generate diverse and customizable content, such as game levels, characters, environments, or narratives, with the goal of enhancing variety, replayability, and user experience \cite{shaker2016procedural}. Previous research in this area \cite{shaker2016procedural, yannakakis2018artificial} relied on evolutionary computation \cite{browne2010evolutionary}, solver-based methods \cite{smith2011answer}, or constructive generation methods (such as cellular automata \cite{johnson2010cellular} and grammar-based \cite{font2016constrained} methods). Deep learning for PCG research \cite{jialin2020yannakakis, summerville2018procedural} has emerged in recent years as a promising approach to generating high-quality game content that is both aesthetically pleasing and challenging while reducing the manual effort required from game developers. Even with these advancements, there are still issues to address, such as ensuring that the content is unique, original, enjoyable to play, and simple to control \cite{summerville2018procedural}.

% To address these issues, our research suggests using conditioned language models and reinforcement learning approaches to content generation \cite{chen2021decision, khalifa2020pcgrl}. These approaches are intended to address the aforementioned issues while also demonstrating how PCG can be used to create high-quality, unique, and playable game content. We hope to demonstrate that these techniques can assist game developers in overcoming the limitations of traditional PCG methods and creating engaging, challenging, and innovative content.

\subsection{Deep Learning-based Level Generation}
Variational Autoencoders (VAE), Generative Adversarial Networks (GANs), and Auto-Regressive Models are a few examples of generative models that rely on deep learning. Many of these models have been used in PCG, such as the generation of Zelda levels using GANs \cite{torrado2020bootstrapping}, the creation and blending of levels from various games using VAEs \cite{sarkar2020conditional}, and the generation of Sokoban levels using LSTMs \cite{suleman2017generation}. When given small datasets, as is frequently the case during game development, these models struggle to generalize because they typically rely on training to capture the distribution of the training data. The bootstrapping method \cite{torrado2020bootstrapping} looks for playable levels in the model's output that can be used to train the model in subsequent iterations to solve this problem. This approach can be used with other generative models in addition to Self-Attention GANs \cite{zhang2019self}. Conditional embedding was used in conjunction with the bootstrapping method \cite{torrado2020bootstrapping}, which increased the generator's quality and diversity. This paper also shows how the generator output can be controlled by conditional embedding. In addition, Generative Playing Networks (GPN) \cite{bontrager2021learning} demonstrated how a generator could be trained to create levels from scratch using input from an agent that had been trained to play the game. By using back-propagating gradients from a critic network, the generator is trained to design difficult yet controllable levels. In order to train without data, the agent is trained on the output of the generator, and the generator is trained on the performance of the agent, resulting in a closed loop where both the agent and the generator get better at the same time. An initial set of levels can be used to train the playing agent, which can then be used to bootstrap the training of the generator.

Training an agent to create levels is another PCG strategy. A level's tiles can be changed using procedural content generation via reinforcement learning (PCGRL) \cite{khalifa2020pcgrl}, which trains an agent to do so and rewards it if the level's quality increases. The controls that allow the user to specify the desired properties of the created level were added to this technique \cite{earle2021learning}. A reward function is used in both strategies to distinguish between low- and high-quality levels. In Adversarial Reinforcement Learning for PCG (ARLPCG) \cite{gisslen2021adversarial}, a generator agent is likewise trained, but the reward is based on how well a playing agent that has been trained to succeed at the generated levels performs. It has been tried out in 3D racing and platforming.

Another recent work uses a transformer-based language model to generate levels for Super Mario Bros in a controlled way \cite{todd2023level}. However, This work is different from ours and targets a different domain, which generates prompt-conditioned content for a side-scrolling game.

\subsection{Sequence Modelling and Transformers}
In the past, sequence modeling has been accomplished using recurrent neural networks (RNNs) \cite{rumelhart1985learning} and long short-term memory (LSTM) networks \cite{hochreiter1997long}. Because of the temporal interdependency of actions, these approaches are constrained in their ability to scale and suffer from the issue of the network's state vector fading with time. 
% Transformers \cite{vaswani2017attention}, on the other hand, get around these problems by using associative attention \cite{bahdanau2014neural} -also referred to as self-attention- to learn reprojections of the input sequence. 
Transformers can efficiently extract the most crucial information from the input sequence window to forecast the following output element by leveraging self-attention module \cite{bahdanau2014neural}. Additionally, the entire input window is presented to the Transformer, allowing for parallelization of operations within a single self-attention layer and scalability of the architecture.

These architectural advancements have paved the way for the development of Large Language Models (LLMs) capable of learning from extensive datasets. Moreover, these models have demonstrated their effectiveness in accelerating the learning of downstream tasks \cite{wei2021pretrained}. Also, the process of fine-tuning LLMs involves initializing model weights with pre-trained values before adapting them to new tasks. Notably, fine-tuning bidirectional LLMs \cite{devlin2018bert, sanh2019distilbert} and unimodal LLMs \cite{radford2019language} has yielded impressive results in tasks involving mask prediction and sequence modeling, respectively.
% Large Language Models (LLMs) can now learn from enormous datasets and speed up the learning of downstream tasks thanks to developments in architecture. Fine-tuning LLMs \cite{devlin2018bert}, which use pre-trained weights for new tasks to initialize the model weights, is one efficient method. Both unimodal LLMs \cite{radford2019language} and bidirectional LLMs \cite{devlin2018bert, sanh2019distilbert} have demonstrated outstanding results on tasks involving mask prediction and sequence modeling.

\subsection{Offline reinforcement learning}
Offline Reinforcement Learning (Offline RL) is an approach to learning from pre-collected data without direct interaction with the environment, as proposed by Levine et al. in 2020 \cite{levine2020offline}. This method relies on a fixed dataset of experiences that were previously gathered by a behavior policy. Offline RL proves particularly advantageous in scenarios where engaging in real-time interactions with the environment is prohibitively costly or poses risks, as observed in fields such as robotics \cite{singh2022reinforcement}, education \cite{singla2021reinforcement}, healthcare \cite{liu2020reinforcement}, and autonomous driving \cite{kiran2021deep}. Moreover, it has been shown to enhance generalization in intricate domains \cite{levine2020offline}.

Unlike online and off-policy RL approaches, which continually interact with the environment to adjust their policies, offline RL operates by deriving a policy solely from a static dataset. Subsequently, once an offline policy is acquired, there exists the option to further refine it through online fine-tuning, which can be a more secure and cost-effective alternative compared to initializing from a random policy, as highlighted by Lee et al. in 2022 \cite{lee2022offline}.

% A major problem in offline RL is the distributional shift: the learned policy and the behavior policy \cite{fujimoto2019off} have different distributions, which can lead to wrong value backups. Previous methods have tried to solve this problem by keeping the learned policy close to the behavior policy using policy regularization \cite{liu2020provably, wu2019behavior, kumar2019stabilizing, zhou2021plas, ghasemipour2021emaq, fujimoto2021minimalist}, conservative value functions \cite{kumar2020conservative}, or model-based training with conservative penalties \cite{yu2020mopo, kidambi2020morel, swazinna2021overcoming, lee2021representation, yu2021combo}.

\begin{figure*}[!t]
  \begin{center}
    \includegraphics[width=\textwidth]{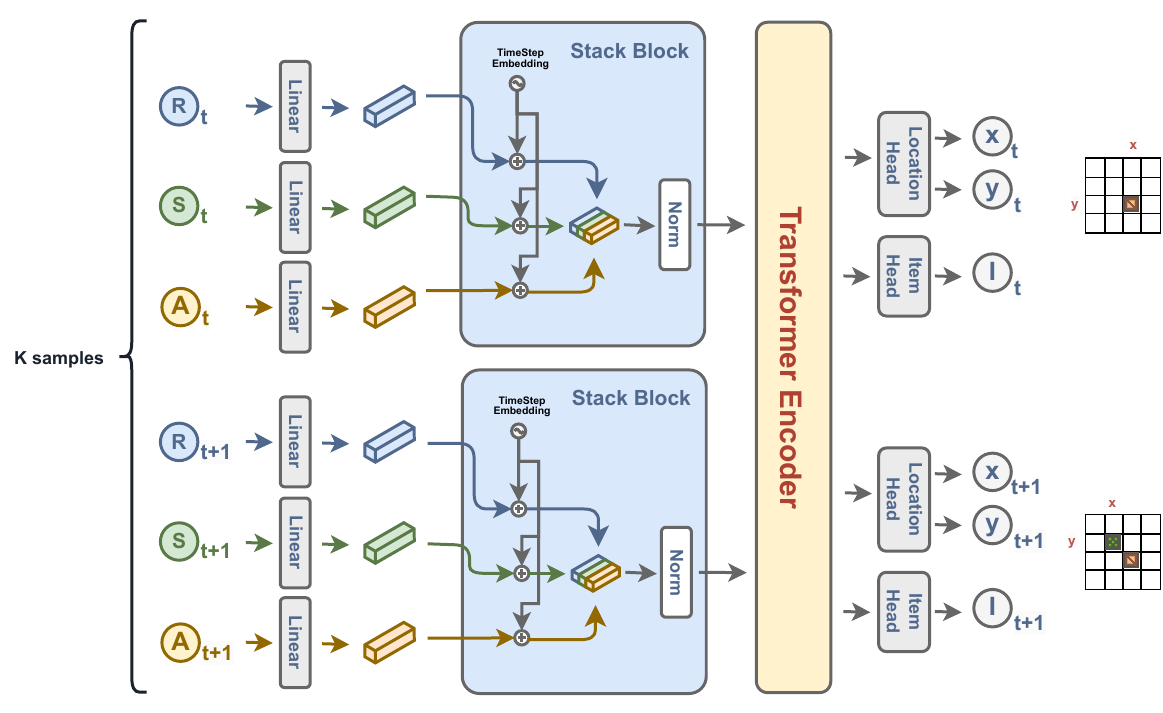}
  \end{center}
  \caption{\small Overview of PCGPT model and Stack Block.}
  \label{Golden_Fig}
\end{figure*}

\begin{equation}
\label{return_to_go_eq.}
rtg = \sum_{t'=t}^{T} r_{(t')}
\end{equation}

\section{Methodology}
This section discusses the PCGPT framework, which presents a novel approach to addressing the PCG problem through the integration of reinforcement learning and transformer networks. The PCGPT Framework adopts an iterative approach to the procedural content generation process, as opposed to creating the complete content in a single iteration. Therefore, content generation can be conceptualized as a Markov Decision Process (MDP), wherein the agent receives an observation and reward at each sequential step and thereafter selects an action in response. In this work, the action is composed of two elements: Item and Location. Item refers to the cell type, and location demonstrates the corresponding position the predicted item should be applied to. This study focuses on the task of level generation, while the concepts discussed below can be extended to other forms of content generation.

\subsection{Content Generation as Classification Problem}
This work examines the PCG as a classification problem. In this regard, we utilize an encoder Transformer architecture \cite{vaswani2017attention}. At each time step, the model determines the optimal action based on prior inputs in four consecutive steps:
\begin{itemize}
    \item Encoding each trajectory in each timestep 
    \item Aggregate the information with self-attention in a transformer encoder
    \item Classifying the optimal action using the item and location heads.
    \item Apply the results to the current trajectory, obtain new state, and repeat the process with the obtained trajectory.
\end{itemize}
The process of iterative content generation begins by establishing an initial state, wherein the map is filled with tiles that are randomly generated.  The agent is only permitted to change one tile at each stage.

PCGPT is different from traditional RL structures since it models trajectories in an autoregressive manner. Some of the main differences between PCGPT and traditional RL structures are:
\begin{itemize}
    \item PCGPT does not require a separate reward function or a value function to guide the agent’s learning. Instead, it uses a causal language model (CLM) objective that maximizes the likelihood of the observed trajectories given the actions taken by the agent.
    \item PCGPT does not use an explicit policy network or a policy gradient algorithm to update the agent’s parameters. Instead, it uses a gradient-based optimization method that minimizes the Negative Log Likelihood loss between the generated and the observed trajectories.
    \item PCGPT does not need state transitions or environment dynamics in the training phase. Instead, It predicts outputs for each time step independently.
    \item PCGPT uses the transformer’s self-attention mechanism to capture the temporal dependencies and the causal relationships between actions and observations.
\end{itemize}

\subsection{Trajectory Representation}
We propose a method for representing the agent’s trajectories in the game environment as sequences of tokens that can be processed by a transformer model. A trajectory is made up of a sequence of the actions (\(A\)), states (\(S\)), and rewards (\(R\)) that the agent goes through as they interact with the game. Our goal is to generate trajectories that achieve a desired level of performance, measured by the return-to-go, which is the sum of future rewards from a given time step (equation \ref{return_to_go_eq.}).  

Where \(r_{(t')}\) is the reward received at time step \(t'\). The reward can be either positive or negative, depending on the outcome of the action. 
In the context of the Sokoban game, a positive reward scenario manifests when Procedural Content Generation (PCG) successfully generates levels wherein the quantity of boxes aligns with the number of targets, thereby engendering a balanced and tractable puzzle. Conversely, a negative reward scenario materializes when PCG encounters difficulty in producing coherent levels, leading to the emergence of fragmented regions within the game environment. These disjointed areas have the potential to render the game either unsolvable or excessively challenging.
% For example, collecting a coin or defeating an enemy can result in a positive reward, while losing a life or falling into a pit can result in a negative reward. 
This way, we obtain a trajectory representation that is well-suited for autoregressive training and generation with transformers.

\subsection{PCGPT Architecture}
PCGPT is composed of some embedding layers, Stack block, one causal transformer, and location and item heads to predict the desired outputs. In order to process the trajectories, we use some embedding layers and encode them into one-dimensional tokens. For each time step, we have tokens corresponding to the three modalities of return-to-go, state, and action, depicted in Figure~\ref{Golden_Fig}

\subsubsection{Embedding Layers and Stack Block}
Like the vanilla transformer \cite{vaswani2017attention}, we use linear layers to embed each of these three modalities to the corresponding embedding vector. A shared-weight linear layer is employed for each type of modality. To elaborate, the embedding layers associated with States share their weights, as do the embedding layers for other modalities. In addition to these embedding layers for each modality, output tokens are passed to StackBlock. As the transformer is permutation invariant, it first adds a learned time step embedding to each token to distinguish what belongs to each time step.  It's worth noting that these embedding vectors differs from the conventional positional embedding used in conventional transformers \cite{vaswani2017attention,ViT}, as one time step in our method corresponds to three tokens. The tokens representations are then stacked together and generate a unique input for each time step. We next apply layer normalization to produce final representations. 

\subsubsection{Transformer Encoder}
After getting token representations, we pass them through PCGPT framework for item and location prediction. our framework comprises a GPT model , which is trained in an autoregressive manner, and helps it predict future actions depending on prior tokens. This procedure enables the model to produce meaningful and coherent behaviors in response to the input trajectory.

\subsubsection{Location and Item Heads}
 As said before, we need to generate the horizontal and vertical positions associated with the predicted item. To this end, we uses two location heads to predict \(loc_x\) and \(loc_y\). These localized coordinates precisely denote the grid cell's position necessitating modification in every time step. On the other hand, the item head specifies the type of cell that the location head has chosen. For example, it could be block, agent, target, box, or empty for Sokoban. If the replaced action is the same as the previous tile content, it is similar to the \(no\_action\) operation.

\subsubsection{Training}
To begin, we constructed a dataset consisting of offline trajectories. From this dataset, we sampled mini-batches of sequence length K. For trajectories shorter than K, we padded dummy samples by repeating the modalities of the last time step. since PCGPT uses a causal transformer, we design proper masks that in each time step, tokens are only allowed to attend to tokens originating from the previous time steps. Moreover, the action token of the current time step is masked. Then, we utilized a Negative-log likelihood loss to train both prediction heads for discrete item and location generation. The losses of each time step are then averaged for backpropagation.

\subsubsection{Inference}
The inference phase starts with giving a randomly created map as an initial token. Alongside the prediction of position coordinates (\(loc_x\) and \(loc_y\)), the corresponding item is also executed. Then, the environment agent calculates the reward and subtracts it from the target reward \(rtg\) (equation~\ref{sub_Rtg}) to prepare it for the next time. For the next time step, prior tokens are concatenated with the newly generated state and \(rtg\), and fed to the model to predict the following action. This operation continues until the obtained state and reward fulfill the stopping criteria.

\begin{equation}
\label{sub_Rtg}
rtg_{(t+1)} = rtg_{(t)} - r_{(t)}
\end{equation}

\section{Experiments}
In this study, we demonstrate the consistent integration of our PCGPT architecture into an OpenAI Gym interface (Brockman et al. (2016)). The integration of this system guarantees compatibility with pre-existing agents, facilitating seamless adaptation and evaluation.

In order to evaluate the efficacy and performance of our system, we performed evaluations on the Sokoban game. The way that we generate rewards for each action plays a crucial role. The rewarding system provides motivations for activities that guide the agent toward the final objective of the challenge while penalizing any deviations from the desired path.

Throughout the evaluation process, the generation process are considered complete once the agent successfully achieves the specified goals. These stopping criteria ensures that our training successfully prepares agents to function at their best and produce the intended results.

\subsection{Sokoban}
Sokoban, a renowned puzzle game developed by Thinking Rabbit in 1982, presents players with the task of navigating a warehouse environment and strategically moving boxes to designated targets, all while ensuring they do not become trapped near to walls. The successful completion of this game necessitates thoughtful strategizing and astute decision-making. It challenges the mind and rewards the player's skill by making well-designed levels that offer the right amount of challenge and solvability.

In order to design a 2D level for the game Sokoban, it is necessary to stick to a set of guidelines. The level design should consist of a single player and an equal quantity of boxes and target areas, all of which are accessible to the player, which guarantees equitable and pleasant gameplay.

A Sokoban solver is employed to maintain a balance between the levels' difficulty, ensuring they are neither excessively simple nor excessively challenging. The solution employs a combination of tree search methods, including Breadth-First Search (BFS) and A*, with a node limit of around 5000. By implementing this intelligent approach, it is ensured that players are able to successfully navigate through each level with a maximum of 18 steps (X = 18).

\subsection{Offline Dataset}

In order to train our proposed PCGPT model, we require a large and varied dataset of PCG maps produced by existing methods. However, obtaining such a dataset is challenging and time-consuming, as most of the available PCG datasets are either small, domain-specific, or not publicly accessible. Therefore, we leverage the PCGRL framework \cite{khalifa2020pcgrl}, which enables the use of reinforcement learning to generate procedural content for various games, such as Sokoban. The framework is flexible and can accommodate different reward functions that define the quality and diversity of the generated content.

We use the pretrained PCGRL models to generate maps for Sokoban. For this game, we generated 3,000 maps with different difficulties. We store each map as a two-dimensional array of integers representing different tiles or objects. We also store the corresponding reward values, items and locations for each map. We made our dataset publicly available for other researchers to use and compare with our PCGPT model.

\begin{figure}[!t]
  \begin{center}
    \includegraphics[width=2.5in]{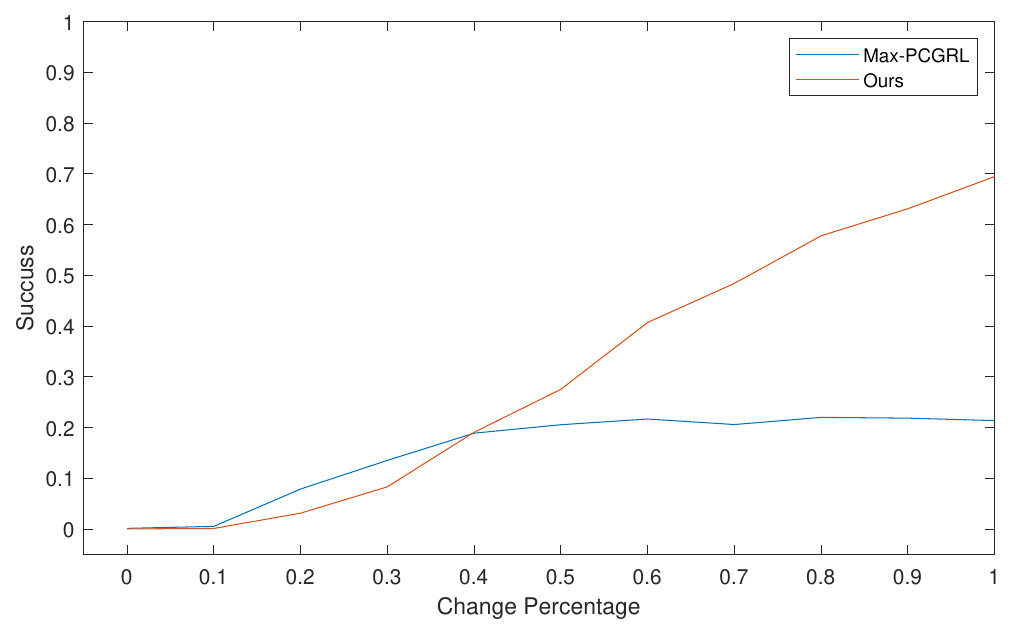}
  \end{center}
  \caption{\small Comparison of success rate between proposed model (PCGPT) and PCGRL for different change percentage values.}
  \label{Change_percentage}
\end{figure}

\subsection{Implementation Details}
The model is trained for 300 epochs with 10000 steps in each epoch. The training procedure required a total of 36 hours on a GPU NVIDIA RTX 3090 in order to successfully train a batch size of 256 samples. The AdamW optimizer was utilized, with a weight decay of 1e-4. In order to prevent the issue of quick convergence during the initial stages of the training process, warm-up steps were implemented. The initial learning rate is set to 1e-4 and subsequently incremented until it reached a value of 1, at which point it remained constant. In order to prevent the issue of overfitting, we incorporated the dropout technique with a probability of 0.1 into the model.

\begin{table}[h!] 
\renewcommand{\arraystretch}{1.3}
\caption{\small Comparison of evaluation factors between PCGPT and PCGRL} 
\label{tab:benchmark results}
\centering
\resizebox{\columnwidth}{!}{%
\begin{tabular}{|c c c c c|}
\hline 
Method & Solution Length & Total Reward & Step & Change \\ 
\hline 
PCGPT & 20.85 & 33.93 & 24.00 & 15.62 \\
\hline 
PCGRL \cite{khalifa2020pcgrl} & 13.35 & 26.82 & 471.98 & 8.18 \\
\hline 
\end{tabular} 
}
\end{table}

\begin{figure*}[!t]
  \begin{center}
    \includegraphics[width=\textwidth]{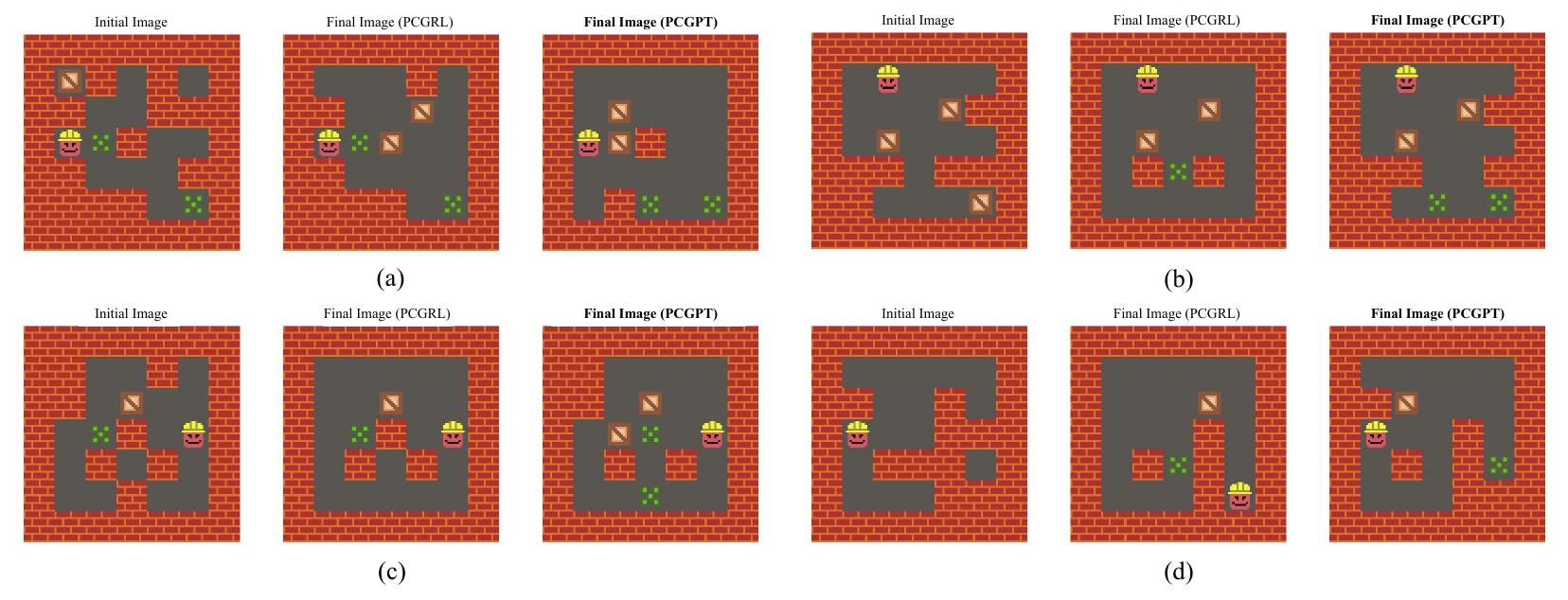}
  \end{center}
  % \caption{\small In (a), our model, PCGPT, generates maps with longer solution lengths, indicating more complexity compared to PCGRL. (b) shows that PCGPT can solve maps that PCGRL cannot. In (c), PCGRL takes 625 iterations with 7 changes, while PCGPT only needs 5 iterations and 5 changes, demonstrating PCGPT's efficiency. Additionally, in some maps like (c) and (d), PCGPT creates more complex but solvable maps by incorporating additional elements or using challenging areas}
    \caption{\small (a): In the generated map, PCGPT produces longer solution lengths, indicating more complex maps compared to PCGRL.  (b): Some PCGRL-generated maps are unsolvable, while PCGPT can generate solvable maps from the same initial state. 
    (c): PCGRL generates a map after 625 iterations with 7 changes, whereas PCGPT generates its map in just 5 iterations with 5 changes, highlighting the efficiency of PCGPT. 
    (d): While both PCGPT and PCGRL create solvable maps, PCGPT generates more complex ones with additional boxes or utilizing the map's challenging aspects, increasing the difficulty level.}
  \label{output_maps}
\end{figure*}

\subsection{Evaluation Results}
For the aim of examining the trained models, we gather 10,000 initial maps that are randomly constructed. These maps are grouped into ten groups of 1,000 maps each. The trained models are then assigned the task of transforming these random initials into solvable game maps. To study the models' capability, we examined them by ranging the change percentage value from 0\% to 100\%. This range represents how much the agent is allowed to modify the tiles during the evaluation. At each stage of change, we test how well the models perform across all ten sets of the dataset.
Figure \ref{Change_percentage} illustrates the ratio of successfully generated maps due to stopping criteria. The horizontal axis reflects the change percentage during inference while the vertical axis represents the proportion of successful levels. PCGRL trained its model using three policies: narrow, turtle, and wide. For the sake of simplicity, here we showed the maximum value of results obtained from each of these policies for the PCGRL model.

Increasing the change percentage let the model to alter more tiles when generating the result map. We can demonstrate that for change percentage values bigger than 0.4, our suggested model beats the PCGRL in all representations i.e. \textit{Narrow}, \textit{Turtle}, and \textit{Wide}. It also shows that the PCGRL model’s curve get saturated and it means that it needs only about 40\% of tiles to build the final map. On the other hand, PCGPT’s curve is monotonically increasing, which shows its potential in utilizing all tiles for building better maps with higher success rates. This is especially intriguing as we used the offline dataset which is created from the PCGRL model which is trained with change percentage of 0.2. We can also see that the PCGPT’s curve is a convex function for change percentages below 0.6 and concave above that. The finding implies that supplying more degree of freedom to the model in early steps of change percentage is more important.

We also compared our proposed approach with the PCGRL from various perspectives and the results are provided in Table 1. In this evaluation we measured solution length, total reward, steps, and changes for all generated solvable maps by each of the methods.

\begin{itemize}
    \item The concept of "Solution Length" refers to the least number of steps required for a player to successfully complete the game. According to the results, PCGPT demonstrates superior performance compared to PCGRL by producing maps of greater complexity that necessitate a greater number of moves to be resolved. So, our model has the capability to generate intricate and varied maps in contrast to PCGRL.
    \item The "Total Reward" refers to the cumulative sum of rewards that are given to the model based on the actions it predicts while completing the output map. According to the data presented in Table 1, the total reward achieved by PCGPT demonstrates a superior performance, approximately 6\% increase, compared to PCGRL. According to this data, PCGPT predicts, on average, more helpful actions in each time step than PCGRL.
    \item The term "Step" in the context of a game refers to the discrete units of time required to generate the map based on the initial input. Table \ref{tab:benchmark results} illustrates a notable reduction in this factor, resulting in a higher throughput during the testing phase.
    \item "Change" quantifies the number of tiles that undergo modification from the original map tiles (25 for Sokoban game) in order to achieve the final map. One intriguing aspect of this term is that our model is capable of modifying a greater number of tiles within a fewer amount of steps. This implies that PCGPT produces a lower number of \({no\_actions}\) compared to PCGRL.
\end{itemize}

The results obtained from running a trained model on the Sokoban game are shown in Figure~\ref{output_maps}. In the conducted experiments, the initial state is kept fixed while our trained model and PCGRL are run for each map, using identical initial maps for both models. The PCGRL model was executed with a change percentage of 100\%, compared to the value of 20\% that authors in \cite{khalifa2020pcgrl} applied during the training phase. The reason behind the using higher values of the change percentage is to enable the algorithm make a greater number of modifications, particularly in cases where it initiates from an undesirable state. Our model used an offline dataset -which is trained by 20\% of change percentage- and is able to change 100\% of tiles during inference.

First, in all of the generated maps the solution length of PCGPT map is longer, which indicates that our model makes more complicated maps compared to PCGRL. Second, we observed that some generated maps by PCGRL are unsolvable, but the PCGPT could generate a solvable map from the same initial. Third, for the sample of Figure~\ref{output_maps}c. it is worth noting that PCGRL generated the map after 625 iterations and with 7 changes in total. While the PCGPT generated its own map in just 5 iterations and 5 changes in total. This observation shows that PCGRL generated lots of \({no\_actions}\), in this case 618, which are redundant and time-consuming. But on the other hand, PCGPT made more accurate and useful actions that could finish generating the final procedure map in 125 times fewer iterations compared to PCGRL. We can also see that in some maps, like Figure~\ref{output_maps}c and Figure~\ref{output_maps}d, both maps are solvable but the generated map by the PCGPT is more complicated which has more boxes or uses the hard part of the map to make it more difficult to solve.

\section{Conclusion}

In conclusion, we proposed a novel framework called PCGPT for procedural content generation using transformers. We demonstrated that PCGPT can generate complex and diverse game maps in the Sokoban game. We compared PCGPT with PCGRL, a state-of-the-art procedural content generation method, and showed that PCGPT outperforms PCGRL in terms of solution length, total reward, steps, and changes. Our approach enables generating high-quality game maps that challenge players and provide diverse gameplay experiences. Overall, our work contributes to the advancement of procedural content generation in video games using deep learning techniques. We hope that this paradigm will foster future work in PCG.

\bibliographystyle{abbrv}
\bibliography{my_bib_file}
\end{document}